\documentclass[11pt]{article}

\usepackage{acl}
\usepackage{makecell}
\usepackage{subcaption}
\usepackage{times}
\usepackage{latexsym}

\usepackage[T1]{fontenc}

\usepackage[utf8]{inputenc}

\usepackage{microtype}

\usepackage{inconsolata}

\usepackage{graphicx}

\usepackage{hyperref}

\title{EXL Health AI Lab at MEDIQA-OE 2025: Evaluating Prompting Strategies with MedGemma for Medical Order Extraction}

\author{
\textbf{Abhinand Balachandran, Bavana Durgapraveen} \\ 
\textbf{Gowsikkan Sikkan Sudhagar, Vidhya Varshany J S, Sriram Rajkumar} \\
EXL Service \\
\texttt{\{abhinand.b, sriram.rajkumar, bavana.durgapraveen\}@exlservice.com}
}

\begin{document}
\nolinenumbers
\maketitle
\pagestyle{empty} 

\begin{abstract}
The accurate extraction of medical orders from doctor-patient conversations is a critical task for reducing clinical documentation burdens and ensuring patient safety. This paper details our team submission to the MEDIQA-OE-2025 Shared Task. We investigate the performance of MedGemma, a new domain-specific open-source language model, for structured order extraction. We systematically evaluate three distinct prompting paradigms: a straightforward one-Shot approach, a reasoning-focused ReAct framework, and a multi-step agentic workflow. Our experiments reveal that while more complex frameworks like ReAct and agentic flows are powerful, the simpler one-shot prompting method achieved the highest performance on the official validation set. We posit that on manually annotated transcripts, complex reasoning chains can lead to "overthinking" and introduce noise, making a direct approach more robust and efficient. Our work provides valuable insights into selecting appropriate prompting strategies for clinical information extraction in varied data conditions.
\end{abstract}

\section{Introduction}

The proliferation of ambient clinical intelligence (ACI) systems promises to revolutionize healthcare by automating the burdensome task of clinical documentation \cite{yim-etal-2023-aci-bench}. A cornerstone of this automation is the ability to transform unstructured doctor-patient dialogue into structured, actionable data suitable for Electronic Health Records (EHRs) \cite{zhang-etal-2023-metadata}. Among the most critical data to capture are medical orders—medications, lab tests, imaging studies, and follow-ups—where accuracy is paramount for patient care and safety \cite{singhal-etal-2023-llms}. The MEDIQA-OE 2025 shared task provides a crucial benchmark for this challenge, pushing the field to develop systems that can parse long, complex conversations to extract a variety of order types and their corresponding clinical justifications \cite{corbeil-etal-2025-mediqa-oe}. This task moves beyond simple entity recognition, requiring a deep understanding of context, negation, and the relationships between a medical order and its underlying reason.

\section{Shared Task and Dataset}

The MEDIQA-OE 2025 shared task \cite{corbeil-etal-2025-mediqa-oe} represents a significant advancement in clinical natural language processing, requiring participants to extract structured medical orders from dialogue transcripts. The dataset, SIMORD \cite{corbeil-etal-2025-simord}, is derived from mock clinical consultations and annotated by medical professionals, providing a robust foundation for developing automated clinical documentation systems.
Medical order extraction involves identifying and structuring various medical orders—such as medications, imaging studies, lab tests, and follow-ups—based on doctor-patient conversations. This complex task goes beyond simple entity recognition, requiring systems to understand the clinical context, temporal relationships, and hierarchical structure of medical directives. Previous efforts in this domain have primarily focused on extracting entities and relations from clinical texts, but have often been limited to structured electronic health records or simplified clinical notes rather than the nuanced, conversational format of real-time clinical interactions.
The MEDIQA-OE 2025 shared task addresses these limitations by presenting a more realistic and challenging scenario that mirrors actual clinical practice. This shared task seeks to develop effective solutions for improving clinical documentation, reducing the administrative burden on healthcare providers, and ensuring that critical patient information is accurately captured from long, complex conversations that may span multiple topics and include interruptions, clarifications, and informal language typical of natural clinical dialogue.
The input dialogues are sourced from a combination of existing conversational datasets, including ACI-Bench \cite{yim-etal-2023-aci-bench}, which focuses on ambient clinical intelligence and automatic visit note generation, and PriMock57 \cite{korfiatis-etal-2022-primock57}, a comprehensive dataset of primary care mock consultations. These datasets provide diverse conversational patterns and clinical scenarios, ensuring that participating systems are evaluated on realistic variations in communication styles, medical specialties, and patient presentations. The structured lists of medical orders are created by qualified medical annotators who possess the clinical expertise necessary to accurately identify, categorize, and structure the complex medical directives that emerge from these conversations, ensuring high-quality ground truth labels that reflect real-world clinical decision-making processes.
This comprehensive approach to dataset creation makes MEDIQA-OE 2025 a valuable benchmark for advancing the state-of-the-art in clinical conversation understanding and automated medical documentation systems. Table 1 presents the distribution of clinical encounters and the corresponding extracted orders in the Train and Dev datasets.
For each conversation, systems must extract all relevant orders and structure them with the following key attributes:
\begin{itemize}
    \item \textbf{Order Type}: The category of the order (e.g., \emph{Medication, Lab, Imaging, Follow-up}).
    \item \textbf{Description}: The specific details of the order (e.g., \emph{``Lisinopril 10mg daily''}).
    \item \textbf{Reason}: The clinical justification for the order (e.g., \emph{``for high blood pressure''}).
    \item \textbf{Provenance}: The specific text spans in the transcript from which the information was extracted.
\end{itemize}

\begin{table*}[htbp!]
  \centering
  \begin{tabular}{lcccccc}
    \hline
    \textbf{Dataset} & \textbf{Encounters} & \textbf{Follow-up} & \textbf{Imaging} & \textbf{Lab} & \textbf{Medication} & \textbf{Total Orders} \\
    \hline
    Train & 63  & 25 & 14 & 29  & 75  & 143 \\
    Dev   & 100 & 41 & 26 & 71  & 117 & 255 \\
    \hline
  \end{tabular}
  \caption{Distribution of clinical encounters and extracted orders across different categories (Follow-up, Imaging, Lab, and Medication) in the Train and Dev datasets.}
  \label{tab:dataset_distribution}
\end{table*}

\section{Related Work}

Clinical Natural Language Processing (NLP) has undergone a significant methodological shift, evolving from rule-based systems to advanced Agentic systems powered by transformers. The Dialogue Medical Information Extraction task was initially addressed by combining Named Entity Recognition (NER) and Relation Extraction (RE). Early rule-based systems relied on semantic lexicons and regular expressions for pattern matching, offering interpretability but facing limitations in scalability and coverage \cite{benabacha-etal-2021-medicqa}.

More recently, supervised heterogeneous graph-based approaches have demonstrated superior performance in mapping medical items to their statuses by enriching their representation with broader dialogue context \cite{zhang-etal-2023-metadata}. Concurrently, GPT-based models utilizing various prompting strategies have been effectively employed for clinical information extraction \cite{lehman-etal-2023-gpt4}

However, much of this prior work has focused on information extraction with minimal emphasis on complex relation identification \cite{yim-etal-2023-aci-bench}. The current challenge extends beyond just medication extraction to encompass lab orders, imaging studies, and follow-up instructions—areas that lack systematic research. A key difficulty lies in accurately mapping orders to their precise reasons, which is crucial for healthcare workflows \cite{sinsky-etal-2024-inbox}. Our research contributes to this area by systematically comparing prompting strategies—from simple in-context learning to complex agentic AI—to develop a robust medical order extraction system for challenging clinical settings \cite{gao-etal-2023-scaling,lewis-etal-2020-rag}.

\section{Methodology}
Our entire approach is built upon the MedGemma family of models, which are variants of Google's Gemma models further pre-trained and fine-tuned on a vast corpus of medical literature and clinical data. This domain-specific tuning endows them with a strong baseline understanding of medical terminology and concepts. We explored both the 4B and 27B parameter variants to assess the impact of model scale. We designed and tested three distinct prompting frameworks.

\subsection{Approach 1: 1-Shot Prompting}

This is our simplest and most direct approach. The model is given a single, high-quality example of a complete conversation transcript and its corresponding structured JSON output. The test transcript is then appended, and the model is instructed to generate the JSON output in the same format. The prompt is structured to be clear and concise, minimizing cognitive load and relying on the model's powerful in-context learning ability to replicate the task.

\subsection{Approach 2: ReAct Framework}
 
Inspired by the ReAct (Reasoning and Acting) paradigm, this approach encourages the model to "think out loud." The process begins with the conversation transcript, which is combined with a system prompt that specifies the extraction categories (medication, lab, imaging, follow-up), rules, and output format. The transcript is then preprocessed by segmenting individual turns, identifying the physician’s utterances, and assigning normalized turn identifiers.
Within each conversation, the extraction follows an iterative ReAct cycle. In the Thought step, the model analyzes the physician’s turns to detect potential medical orders. The Action step then generates candidate extractions, specifying the order type, a short description, the associated clinical reason, and the provenance (turn numbers). The Observation step validates these candidates by enforcing constraints: only doctor-initiated orders are kept, exact transcript wording must be preserved, descriptions and reasons are limited to 20 words, compound instructions are split, duplicate orders are removed, and the number of orders and provenance entries are capped. When inconsistencies are identified, the cycle repeats until valid outputs are produced.
After this reasoning loop, a post-processing stage ensures consistency by normalizing order types, truncating fields to required lengths, ordering entries by their provenance, and validating against the target JSON schema. The final system output is a structured JSON array containing all extracted medical orders, with each entry including order type, description, reason, and provenance.
This method aims to improve accuracy on complex cases by forcing the model to explicitly reason about its decisions before producing the final output.




\begin{figure*}[t]
  \centering
  \begin{subfigure}{0.9\linewidth}
    \includegraphics[width=\linewidth]{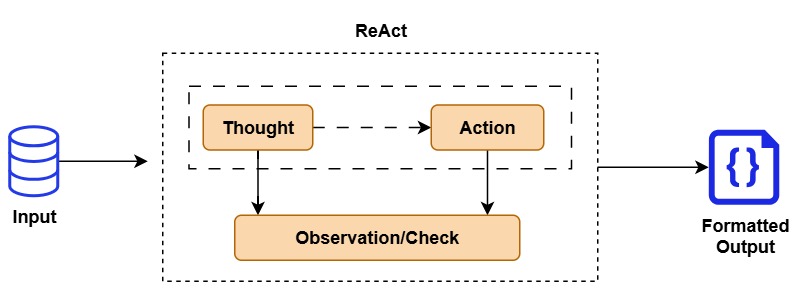}
    \caption{React Approach Flow}
    \label{fig:desc_image}
  \end{subfigure}

  \vspace{0.5cm} 

  \begin{subfigure}{0.9\linewidth}
    \includegraphics[width=\linewidth]{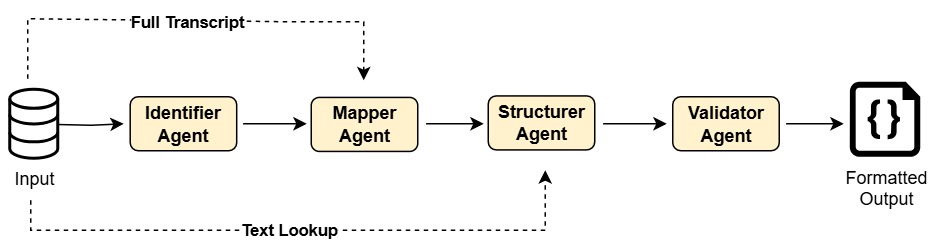}
    \caption{Agentic Approach Flow}
    \label{fig:other_image}
  \end{subfigure}

  \caption{Overview of ReAct and Agentic Workflow frameworks for medical order extraction. The ReAct approach (Figure a) implements iterative thought-action-observation cycles where the model explicitly reasons through extraction decisions, validates candidates against constraints, and repeats until consistent outputs are achieved. The Agentic Workflow (Figure b) decomposes the task across four specialized agents: Identifier for detection, Mapper for pairing orders with reasons, Structurer for JSON formatting, and Validator for final verification.}
  \label{fig:vertical_figures}
\end{figure*}

\subsection{Approach 3: Agentic Workflow}
This is our most complex framework, decomposing the task across a simulated multi-agent pipeline within a single prompt context.
Agent 1 (Identifier): Scans the entire transcript turn-by-turn and outputs a raw list of potential orders and reasons, tagged with their turn IDs.
Agent 2 (Mapper): Takes the output from Agent 1. Its sole job is to analyze the raw list and create explicit pairings between each identified order and its most likely reason.
Agent 3 (Structurer): Receives the mapped pairs from Agent 2. It formats this information into the final, clean JSON structure, ensuring all fields are correctly populated.
Agent 4 (Validator): Performs a final check on the generated JSON, comparing it against the original transcript to correct any obvious errors or hallucinations before producing the final output.
This workflow was designed to modularize the cognitive process, hoping to reduce errors by having specialized "agents" focus on one sub-task at a time.

\section{Evaluation metrics}
\label{sec:bibtex}

We assess our system using four evaluation metrics, each designed to capture different dimensions of clinical order extraction. For the assessment of order descriptions and underlying reasons, we employ the ROUGE-1 F1 score which quantifies unigram-level overlap between predicted outputs and gold-standard references. ROUGE-1 is a standard metric in clinical NLP for measuring content accuracy in generated text, making it appropriate for evaluating free-text fields such as descriptive notes and justifications. For order type classification, we adopt a strict F1 metric, in which predictions are counted as correct only if they exactly match the annotated label. This stricter criterion prevents partial matches from inflating scores and is especially important when separating clinically distinct categories like medications, laboratory tests, imaging, and follow-up visits. For the task of provenance detection, we report a multi-label F1 score, since an order can be linked to multiple conversational sources (e.g., several speakers or turns). Multi-label evaluation provides a balanced measure of precision and recall in these cases, offering a more realistic view of performance. Collectively, these metrics evaluate both the textual alignment of generated outputs and the accuracy of structured predictions, resulting in a well-rounded framework for assessing clinical order extraction.

\section{Results and Discussion}

We evaluated our three approaches on the official validation set to systematically assess their performance and determine the optimal strategy for medical order extraction. Our experimental design was structured to answer a fundamental research question: Which prompting strategy demonstrates the highest effectiveness for this specialized clinical task? The evaluation framework encompassed multiple metrics to provide comprehensive insights into model performance across different aspects of the order extraction process.
Initial experiments comparing the base Gemma model with its medical adaptation revealed significant improvements with domain-specific fine-tuning. MedGemma consistently outperformed the base model across all evaluation metrics, demonstrating the value of medical domain adaptation for clinical natural language processing tasks. This finding underscores the importance of specialized model training for healthcare applications, where domain-specific terminology and contextual understanding are crucial for accurate performance. 
Building upon this foundation, we conducted a comprehensive comparison of our three prompting strategies using the MedGemma-4B model on the development dataset. The results, presented in Table 2, revealed a clear performance hierarchy among the approaches. The 1-Shot methodology achieved the highest average score of 0.436, demonstrating superior performance across most evaluation dimensions. Specifically, it excelled in description generation (Rouge1-f1: 0.516) and order type classification (Strict-f1: 0.602), while maintaining competitive performance in reason extraction and provenance identification. In contrast, the ReAct approach yielded significantly lower scores across all metrics (average: 0.277), while the Agentic Workflow showed mixed results, performing well in certain areas like provenance detection (MultiLabel-f1: 0.488) but struggling with reason extraction (Rouge1-f1: 0.276). The results, presented in Table \ref{tab:medgemma27b_results} , revealed that 1-shot using MedGemma 27B approach stood at fourth place in the MEDIQA-OE 2025 shared task \cite{corbeil-etal-2025-mediqa-oe}
\begin{table*}[htbp!]
  \centering
  \begin{tabular}{lccccc}
    \hline
    \textbf{MedGemma-4B} & 
    \makecell{\textbf{Description}\\\textbf{(Rouge1\_F1)}} & 
    \makecell{\textbf{Reason}\\\textbf{(Rouge1\_F1)}} & 
    \makecell{\textbf{Order Type}\\\textbf{(Strict\_F1)}} & 
    \makecell{\textbf{Provenance}\\\textbf{(MultiLabel\_F1)}} & 
    \textbf{Avg. Score} \\
    \hline
    1-Shot           & 0.516 & 0.318 & 0.602 & 0.307 & 0.436 \\
    ReAct            & 0.363 & 0.120 & 0.465 & 0.160 & 0.277 \\
    Agentic Workflow & 0.09 & 0.06 & 0.169 & 0.123 & 0.111 \\
    \hline
  \end{tabular}
  \caption{Comparison of prompting strategies with MedGemma-4B with dev dataset. The 1-Shot approach yielded the best overall performance.}
  \label{tab:medgemma_results}
\end{table*}

\begin{table*}[htbp!]
  \centering
  \begin{tabular}{lccccc}
    \hline
    \textbf{MedGemma-27B} & 
    \makecell{\textbf{Description}\\\textbf{(Rouge1\_F1)}} & 
    \makecell{\textbf{Reason}\\\textbf{(Rouge1\_F1)}} & 
    \makecell{\textbf{Order Type}\\\textbf{(Strict\_F1)}} & 
    \makecell{\textbf{Provenance}\\\textbf{(MultiLabel\_F1)}} & 
    \textbf{Avg. Score} \\
    \hline
    1-Shot & 0.591 & 0.342 & 0.703 & 0.561 & 0.549 \\
    ReAct  & 0.353 & 0.283 & 0.497 & 0.350 & 0.370 \\
    \hline
  \end{tabular}
  \caption{Comparison of prompting strategies with the larger MedGemma-27B model on the test dataset. The 1-Shot approach remains the most effective.}
  \label{tab:medgemma27b_results}
\end{table*}
To validate these findings and assess scalability, we replicated the experimental framework using the larger MedGemma-27B model on the test dataset. The results, shown in Table 3, confirmed our initial observations while demonstrating the benefits of increased model capacity. The 27B model achieved substantially higher scores across all metrics compared to its 4B counterpart, with the 1-Shot approach reaching an average score of 0.549. Notably, the performance improvements were particularly pronounced in description generation (Rouge1-f1: 0.591) and order type classification (Strict-f1: 0.703). Despite the overall performance gains from the larger model, the relative ranking of prompting strategies remained consistent, with 1-Shot maintaining its superiority over the ReAct approach (average: 0.370).
Our analysis revealed a counterintuitive but significant finding: increased complexity in prompting strategies did not translate to improved performance for this specific task. The ReAct and Agentic frameworks, despite their theoretical sophistication and success in other domains, consistently underperformed relative to the simpler 1-Shot approach. Through detailed error analysis, we identified that this phenomenon stems from what we term "analytical over-processing"—the more complex frameworks occasionally generated spurious intermediate reasoning steps that introduced errors rather than enhancing accuracy. The models would sometimes fabricate relationships between dialogue elements or misinterpret subtle clinical nuances during their multi-step reasoning processes, ultimately degrading precision. Since the MEDIQA-OE dataset consists of carefully annotated clinical transcripts with well-defined ground truth, the additional inferential layers introduced by complex prompting strategies contributed more noise than valuable signal. The 1-Shot approach, by maintaining a more direct and constrained generation process, proved less susceptible to such systematic errors while offering additional benefits in terms of computational efficiency and implementation simplicity.

\section*{Limitations}
The primary limitation of our study is tied to our main finding. Our conclusion that 1-Shot prompting is superior is heavily dependent on the MEDIQA-OE dataset. In a real-world clinical setting with noisy ASR transcripts, interruptions, and less structured speech, the explicit reasoning steps of a ReAct or Agentic framework might be necessary to disambiguate the input and could potentially outperform a direct 1-Shot approach. Our work does not test this hypothesis. Furthermore, our Agentic workflow was implemented within a single model using four specialized agents to handle different subtasks; while this design demonstrates feasibility, a true multi-agent system with independent models could behave differently. This multi-agent setup also introduces practical limitations, as larger models such as MedGemma-27B make the Agentic approach computationally expensive and time-consuming, thereby increasing cost. At the same time, smaller models such as MedGemma-4B exhibited notable shortcomings: we observed hallucinations in ReAct and Agentic reasoning steps, as well as a tendency to replicate the few-shot examples provided rather than extracting new information from the transcript.
In addition, there are several dataset-related limitations. Primarily, the available annotations are relatively sparse and at times inconsistent, with inter-annotator variability leading to ambiguity in what counts as a valid order. secondly, the dataset does not always provide medically precise or standardized order labels, which limits the ability to evaluate correctness against clinically meaningful ground truth. Finally, certain order categories are underrepresented, creating class imbalance that could bias model performance.

\section*{Conclusion}
In this paper, we presented our investigation into medical order extraction for the MEDIQA-OE 2025 task. By systematically comparing 1-Shot, ReAct, and Agentic prompting frameworks with the MedGemma model, we demonstrated that for manually annotated clinical transcripts, a direct and simple 1-Shot approach is surprisingly effective. It outperformed more complex reasoning frameworks, which were prone to overthinking and introducing errors. This highlights a crucial lesson for applied NLP: the optimal solution is a function of not just the model's power, but also the characteristics of the data. Future work should explore these prompting paradigms on noisier, real-world clinical data to determine if the utility of complex reasoning frameworks becomes more apparent.

\section*{Acknowledgments}
We extend our sincere thanks to EXL Health AI Lab for their support and computing resources. We also appreciate the efforts of our colleagues who contributed to discussions and provided valuable assistance during the course of this work. Finally, we acknowledge the organizers for their efforts in hosting this interesting and challenging competition.


\bibliography{custom}
\appendix
\section{\textbf{Prompts used in this approach}}
One-shot approach prompt which was in \textbf{fourth place in the MEDIQA- OE 2025 competition} is mentioned below and the remaining experimental approach prompts are available in Git repo

\subsection{\textbf{One-shot approach prompt}}
\textbf{SYSTEM PROMPT}: """
You are a medical AI assistant specialized in extracting EXPLICIT medical orders from doctor-patient conversations.

CRITICAL RULES:
1. Extract ONLY orders explicitly stated by the doctor
2. Do NOT infer or assume orders that aren't clearly mentioned
3. Provenance must be EXACT turn numbers where orders appear
4. Be balanced - i.e precision and recall on level terms
5. If the doctor orders multiple DISTINCT items (e.g., 'get a covid test and blood test'), create separate order objects for each item - never merge them into one combined description.

Order Types:
- medication: Prescriptions, dosage instructions, medication changes
- lab: Blood tests, urine tests, specific diagnostic tests
- imaging: X-rays, MRI, CT scans, ultrasounds
- followup: Scheduled return visits, check-ups (these must be explicitly stated by the doctor)

For each order extract:
- order type: One of the 4 types above
- description: EXACT medical terminology used by doctor
- reason: Specific condition/symptom mentioned by doctor
- provenance: ONLY turn numbers where this exact order is mentioned"""

\textbf{INSTRUCTION TEMPLATE} : """Please extract all medical orders from the following doctor-patient conversation:

CONVERSATION:
{conversation}

Extract all medical orders and return them as a JSON list with the following format:
[
  {{
    "order type": "medication|lab|imaging|followup|referral",
    "description": "specific description of the order",
    "reason": "medical condition or reason for the order", 
    "provenance": [list of turn numbers where this order appears]
  }}
]

Focus on explicit orders given by the doctor. Be precise with medical terminology."""

\textbf{USER PROMPT}:  f"""EXAMPLE CONVERSATION:
Turn 126 - DOCTOR: so, for your first problem of your shortness of breath i think that you are in an acute heart failure exacerbation.
Turn 127 - DOCTOR: i want to go ahead and, uh, put you on some lasix, 40 milligrams a day.
Turn 138 - DOCTOR: for your second problem of your type i diabetes, um, let's go ahead... i wanna order a hemoglobin a1c for, um, uh, just in a, like a month or so.

EXPECTED OUTPUT:
[
   \{ \{
    "order type": "medication",
    "description": "lasix 40 milligrams a day",
    "reason": "shortness of breath acute heart failure exacerbation",
    "provenance": [126, 127]
   \} \},
   \{ \{
    "order type": "lab", 
    "description": "hemoglobin a1c",
    "reason": "type i diabetes",
    "provenance": [138]
   \} \}
]

NOW EXTRACT FROM THIS CONVERSATION:

---

 \{instruction template \}
"""

\end{document}